# Review of Swarm Intelligence-based Feature Selection Methods

**Mehrdad Rostami[1], Kamal Berahmand[2], Saman Forouzandeh[3]**
Department of Computer Engineering, University of Kurdistan, Sanandaj, Iran[1]
Department of Science and Engineering, Queensland University of Technology, Brisbane, Australia[2]
Department of Computer Engineering University of Applied Science and Technology, Center of Tehran Municipality ICT org.Tehran, Iran[3]
M.rostami@eng.uok.ac.ir[1], kamal.berahmand@hdr.qut.edu.au[2], Saman.forouzandeh@gmail.com[3]

## Abstract:
In the past decades, the rapid growth of computer and database technologies has led to the rapid growth of large-scale datasets. On the other hand, data mining applications with high dimensional datasets that require high speed and accuracy are rapidly increasing. An important issue with these applications is the curse of dimensionality, where the number of features is much higher than the number of patterns. One of the dimensionality reduction approaches is feature selection that can increase the accuracy of the data mining task and reduce its computational complexity. The feature selection method aims at selecting a subset of features with the lowest inner similarity and highest relevancy to the target class. It reduces the dimensionality of the data by eliminating irrelevant, redundant, or noisy data. In this paper, a comparative analysis of different feature selection methods is presented, and a general categorization of these methods is performed. Moreover, in this paper, state-of-the-art swarm intelligence- are studied, and the recent feature selection methods based on these algorithms are reviewed. Furthermore, the strengths and weaknesses of the different studied swarm intelligence-based feature selection methods are evaluated.

## 1. Introduction

Pattern recognition is one of the most important applications of machine learning in different sciences. Machine learning methods utilized in the many areas of medical diagnosis (Liu, Gu et al. 2017, Li, Wu et al. 2019), marketing (Cheng-Lung Huang and Tsai 2009, Yuan, Yuan et al. 2020), image processing (Liang, Zhang et al. 2017, Zhou, Gao et al. 2017), text mining (Wang and Hong 2019, Kou, Yang et al. 2020), information retrieval (Chuen-Horng Lin, Huan-Yu Chen et al. 2014, Ji, Shen et al. 2019), Identification (Bi, Suen et al. 2019, Koide, Miura et al. 2020), etc. One of the significant goals of modeling and classification of data is to predict based on the train data and available features. Huge datasets with high dimensional features space and a relatively smaller number of samples are critical issues for machine learning tasks. Once there are a number of irrelevant and redundant features among the initial feature set, dimensionality reduction is one of the essential techniques to eliminate these features. (Chen, Li et al. 2019). It is an efficient technique for improving accuracy performance, lowering computational complexity, building more generalized models, and decreasing the required storage (Tang, Dai et al. 2019, Wang, Zhang et al. 2019). During the past few years, two major techniques have been proposed for the reduction of dimensionality: Feature extraction and feature selection (Ahmed K. Farahat, Ali Ghodsi et al. 2013). In feature extraction, the primary feature space is mapped to a smaller space. In fact, in this technique, by combining existing features, fewer features are created so that these features contain all (or most of) the information contained in the primary features. Moreover, in feature selection, a subset of initial features is selected by removing the irrelevant and redundant feature.

In general, the feature selection process consists of four main stages: subset generation, subset evaluation, stopping criteria, and validation of results. In each iteration of the search process, a subset of the candidate feature set is generated from the original features, and its appropriateness is measured by an evaluation criterion. The subset generation process and its evaluation are repeated until a predetermined stop criterion is reached. At the end of this process, the best subset of the selected feature is validated on the test dataset.

Feature selection has been an active research area in data mining, pattern recognition, and statistics communities (Liu, Nie et al. 2019). The total search space to find the most relevant and non-redundant features, including all possible subsets, is $2^n$, where $n$ is the number of original features. Comprehensive search ensures that the most appropriate features are found, but usually, this is not computationally feasible, even for medium-sized datasets. Since the evaluation of all possible subsets is very costly, a solution must be searched that is both computationally feasible and useful in terms of quality. Many feature selection methods use metaheuristic algorithms to avoid increasing computational complexity (Alshamlan, Badr et al. 2015, Welikala, Fraz et al. 2015, Singh and Singh 2019). These algorithms will be able to optimize the problem of feature selection with appropriate accuracy within an acceptable time. Metaheuristic algorithms can be classified into two main categories: Swarm intelligence (SI) and Evolutionary Algorithms (EA). SI is a relatively new category of evolutionary computation comparing with EAs and other single-solution based approaches. SI algorithms utilized approximate and non-deterministic techniques to effectively and efficiently explore and exploit the search space in order to find near-optimal solutions The most popular nature-inspired meta-heuristic group is swarm-based techniques. Swarm Intelligence (SI) is a type of artificial intelligence that is based on collective behaviors in decentralized and self-organized systems. These systems usually consist of a population of simple actors who interact locally and with their environment.

In recent decades, many SI-based algorithms have been employed to feature section(Figueiredo, Macedo et al. 2019). Despite the highly acceptable performance of these methods, there are only a few papers reviewing SI based feature selection methods. In (Basir and Ahmad 2014) a comparison of swarm intelligence-based feature selection method is conducted. The authors of this paper focused only on the well-known and traditional SI algorithm, and only the feature selection methods based on these algorithms studied. Moreover, in (Brezočnik, Fister et al. 2018) a comprehensive literature review of SI algorithms for feature selection is performed. One of the limitations of this paper is that it lacks experiment results and only states the strengths and weaknesses of different SI-based feature selection methods. With the lack of comprehensive review of SI-based feature selection methods, the main purpose of this paper is to fill the gap in coverage of SI algorithms for feature selection. This paper seeks to provide a comprehensive overview of SI-based feature selection methods and their categorization and try to review the state-of-the-art and most well-known SI-based method used to feature selection. Also, in this paper, various experiments have been designed to compare the performance of different SI-based methods to allow a more accurate evaluation of these algorithms.

The remainder of this paper is organized as follows: Section 2 presents the introduction of the feature selection problem, Section 3 reviews SI based feature selection methods. Section 4 reports the experimental results of different SI-based methods. Finally, Section 5 presents the conclusion.

## 2. Background

An essential issue with machine learning techniques is the high-dimensionality problem of a dataset where the feature subset size much greater than pattern size. For example, in the medical applications that include very high-dimensional datasets, the classification parameters are also increased. Therefore, the performance of the classifier declines significantly (Liu and Yu 2005, Saeys, Inza et al. 2007, Chandrashekar and Sahin 2014). According to a general rule for a classification problem with $n$ dimension and $C$ class, at least $10 \times n \times C$ training data is required (Anil K. Jain, Robert P. W. Duin et al. 2000, Liu and Zheng 2006, Cadenas, Garrido et al. 2013). When it is not possible to provide this number of training data practically, reducing the feature subset size, reduces the number of required training data. As a result, the performance of the classification algorithm increases (Gokalp, Tasci et al. 2020, Shu, Qian et al. 2020).

From a general point of view, feature selection methods are divided into two categories supervised and unsupervised feature selection methods(Tang, Liu et al. 2018). In supervised methods, a set of train data is available, each of which is described by taking features values along with the class label, while in unsupervised methods, and train data lacks class tags. In general, it can be said that feature selection methods have better efficiency and more reliable performance in the supervised mode due to the use of class labels. Therefore, it is more difficult to select a feature in the unsupervised mode, and in many studies, this area has been considered (Zhang, Zhang et al. 2019, Ding, Yang et al. 2020).

Feature selection methods generally search through the solution space to optimize two conflicting objectives: maximizing the relevancy to the target class and minimizing the redundancy of selected features. Many search strategies are employed to optimize these objectives. These methods are generally categorized into single-objective and multi-objective methods (Senawi, Wei et al. 2017, Hu, Gao et al. 2018). In the single-objective methods, the population is optimized using only one objective in the fitness function. As a result, the choice of objective and definition of the fitness function will greatly affect the accuracy of the optimization algorithm. Also, there are usually several objectives in many optimization issues, and defining a fitness function with just one goal reduces optimization performance. One approach to overcome these challenges is to consider is to consider several different objectives in the fitness function of the feature selection problem. Modeling feature selection as a multi-objective problem can obtain a set of non-dominated feature subsets to meet different requirements in real-world applications. Most of the previously proposed feature selection methods are utilized single-objective function, and there are only a few methods of multi-objective function, such as MOPSO (Xue, Zhang et al. 2013), MOGA (Morita, Sabourin et al. 2003), MOFSEF (Wu, Liu et al. 2020), MOFSEEG (Martín-Smith, Ortega et al. 2017) and MOFSGADM (Li, Xue et al. 2020).

Previous methods for select optimal feature sets can be categorized into four groups consist of filter, wrapper, embedded, and hybrid models. Also, in recent years, graph theoretic-based techniques have been used in many feature selection methods. These categories are explained in the following subsections.

## 2.1. Filter model

The filter model uses statistical and probabilistic data properties to calculate the feature relevance and performs the feature selection process independently of machine learning algorithms. In other words, this model uses the inherent properties of data to evaluate features (Labani, Moradi et al. 2018). The filter model based on how features are evaluated is divided into two univariate and multivariate categories. In the univariate feature selection approaches, the relevance of each feature is calculated in accordance with a given measure such as the Information Gain (IG) (Raileanu and Stoffel 2004), Gain Ratio (GR) (Mitchell 1997), Term Variance (TV) (S. Theodoridis and Koutroumbas 2009), Mutual information (MI) (Xu, Jones et al. 2007), Gini Index (GI) (Raileanu and Stoffel 2004), Laplacian score (LS) (Xiaofei He, Deng Cai et al. 2005) and Fisher score (FS) (Quanquan Gu, Zhenhui Li et al. 2011). In these univariate methods, it is assumed that features independent of each other and possible dependencies between features are not considered. This simple default assumption is wrong and, in many cases, may reduce the efficiency of feature selection methods. On the other hand, multivariate methods measure the appropriateness of features with regard to their interdependence. Therefore, multivariate methods have more computational complexity than univariate methods, but they also have higher performance. There are some multivariate methods, including Minimal Redundancy Maximal Relevance (mRMR) (Peng, Long et al. 2005), Relevance redundancy feature selection (RRFS) (Ferreira and Figueiredo 2012), MIFS (Battiti 1994), Normalized Mutual Information Feature Selection (NMIFS) (Estévez, Tesmer et al. 2009), MIFS-U (Kwak and Choi 2002), Unsupervised Feature Selection based on Ant Colony Optimization (UFSACO) (Tabakhi, Moradi et al. 2014), and Multivariate RDC (MRDC) (Labani, Moradi et al. 2018).

## 2.2. Wrapper model

The wrapper-based methods use a classifier to measure the efficiency of the selected feature subset. In this approach, a search method is used to find the optimal feature subset, and at each stage of the searching strategy, a subset of features is generated and measured by a classifier or another learning model. Finally, the best-produced feature subset is selected as the final feature subset.(Liu and Yu 2005, Chandrashekar and Sahin 2014). Although the wrappers model may select a better feature subset, they are expensive to run and can break down with high dimensional medical dataset. This is due to the use of learning algorithms in the feature subset quality calculation.

## 2.3. Hybrid model

The hybrid model attempts to take advantage of both the filter and wrapper models and to provide a model that balances the computational efficiency of the filter model and the accuracy of the wrapper model. In fact, the purpose of this model is to provide a method that is both efficient and effective.

## 2.4. Embedded model

Moreover, in the embedded model, the feature selection process is considered as part of the learning algorithm. In other words, in this model, a learning algorithm is utilized for searching the optimal feature set (Zhang, Wu et al. 2015).

## 2.5. Graph-based methods

Moreover, recently, graph-based methods are used in machine learning techniques to extract the similarity relationships among the data. In feature selection, graph-based methods provide an underlying manifold structure as a universal framework to reflect the relationships between features. Several research efforts employed graph-based methods for solving the feature selection problem. For example, in (Bandyopadhyay, Bhadra et al. 2014), a dense subgraph finding approach is adopted for the unsupervised feature selection problem. Another clustering-based feature subset selection algorithm for high dimensional data is proposed in (Song, Ni et al. 2013). This work utilizes a graph-theoretic clustering method for similar grouping features. In (Zhang and Hancock 2012), a hypergraph-based method is proposed for feature selection. This work uses an information-theoretic criterion to evaluate the appropriateness of different features by considering the related class label of each sample. In (Moradi and Rostami 2015), the concept of graph clustering with the node centrality measure is integrated with the unsupervised feature selection process. This work is extended by the authors of (Ghaemi and Feizi-Derakhshi 2016) to choose more informative features. In (Henni, Mezghani et al. 2018), the authors employed Google's PageRank centrality measure to rank features based on their importance. Hashemi et al. (Hashemi, Dowlatshahi et al. 2020) propose another graph-based feature selection method for the multi-label high dimensional dataset. The authors of this paper utilize the PageRank centrality measure to rank the features based on their value in the graph. Also, in this paper, the correlation distance criterion is used to remove irrelevant features. In (Li, Tang et al. 2019), an unsupervised graph-based feature selection method for high dimensional data is proposed. In this paper, Laplacian graph and local geometrical structure are used for better representation of the features space. In (Zhu, Zhu et al. 2017), a subspace clustering guided unsupervised feature selection method is proposed. This work uses the subspace clustering to learning of the clustering and then those features with good preservation ability of the cluster labels are selected.

In Table 1 the main characteristic of different feature selection methods is summarized. In this table, six attributes (i.e., the number of objectives, Type of method, search strategy, application, weakness, and strength) of different feature selection methods are reported.

**Table 1**: The main characteristics of different feature selection methods. The synonyms used in this Table are: **MOP**: Multi-Objective Optimization, **SOP**: Single-Objective Optimization, **RB**: Ranking-Based, **SSB**: Subset selection-based, **ACO:** Ant Colony Optimization, **GA:** Genetic Algorithm, **ABC:** Artificial Bee Colony, **DE:** Differential Evolution, **NSGA:** Nondominated Sorting Genetic Algorithm

| Methods | Number of Objectives | Type | Search Strategy | Application | Weakness | Strength |
|---|---|---|---|---|---|---|
| **RDC(Rehman, Javed et al. 2015)** | SOP | Filter-RB | Univariate | Textual | Ignoring feature dependency, Low performance | Fast, Independent of classifier |
| **DFS(Uysal and** | SOP | Filter-RB | Univariate | Textual | | |
| **Gunal 2012)** | | | | | | |
| **NDM(Rehman, Javed et al. 2017)** | SOP | Filter-RB | Univariate | Textual | | |
| **FS (Quanquan Gu, Zhenhui Li et al. 2011)** | SOP | Filter-RB | Univariate | Textual/Numeric | | |
| **GI (Raileanu and Stoffel 2004)** | SOP | Filter-RB | Univariate | Textual/Numeric | | |
| **MI (Xu, Jones et al. 2007)** | SOP | Filter-RB | Univariate | Textual/Numeric | | |
| **LS (Xiaofei He, Deng Cai et al. 2005)** | SOP | Filter-RB | Univariate | Numeric | | |
| **IG (Raileanu and Stoffel 2004)** | SOP | Filter-RB | Univariate | Textual/Numeric | | |
| **CHI(Sebastiani 2002)** | SOP | Filter-RB | Univariate | Textual/Numeric | | |
| **GR (Mitchell 1997)** | SOP | Filter-RB | Univariate | Numeric | | |
| **TV (S. Theodoridis and Koutroumbas 2009)** | SOP | Filter-RB | Univariate | Numeric | | |
| **RRFS(Ferreira and Figueiredo 2012)** | SOP | Filter | Multivariate Greedy | Numeric | Using greedy search strategy and easily trapping into the local optima. | Having higher computational complexity than univariate methods |
| **mRMR(Peng, Long et al. 2005)** | SOP | Filter | Multivariate Greedy | Numeric | | |
| **RSM (Carmen Lai, Marcel J.T. Reinders et al. 2006)** | SOP | Filter | Multivariate Greedy | Numeric | | |
| **GUFS (Ahmed K. Farahat, Ali Ghodsi et al. 2013)** | SOP | Filter | Multivariate Greedy | Numeric | | |
| **MECY_FS (Wang, Li et al. 2015)** | MOP | Filter | GA | Numeric | Only applicable to numeric datasets, Having lower performance in text datasets. | Using information theoretic methods to evaluate the objective function that has made them fast. |
| **RRFSACO(Tabakhi and Moradi 2015)** | SOP | Filter/SSB | Multivariate ACO | Numeric | Being single-objective, Forcing the evolving population to form a particular feature set due to the use of a single quality function. | Considering the feature dependency |
| **GCACO(Moradi and Rostami 2015)** | SOP | Filter/SSB | Multivariate ACO | Numeric | | |
| **MGCACO(Ghimatgar, Kazemi et al. 2018)** | SOP | Filter/SSB | Multivariate ACO | Numeric | | |
| **FASTFS (Song, Ni et al. 2013)** | SOP | Filter/SSB | Multivariate Graph-based | Textual/Numeric | | |
| **GCNC (Moradi and Rostami 2015)** | SOP | Filter/SSB | Multivariate Graph-based | Numeric | | |
| **FSGA (Wenzhu Yang, Daoliang Li et al. 2011)** | SOP | Filter/SSB | Multivariate GA | Foreign Fiber | | |
| **MICGSOFS (Lyu, Wan et al. 2017)** | SOP | Filter/SSB | Multivariate Sequential | Biomedical Datasets | | |
| **MBFFS (Hua, Zhou et al. 2020)** | SOP | Filter/SSB | Markov Blanket | Numeric | | |
| **BT-SFS (Yan, Ma et al. 2018)** | SOP | Wrapper | Sequential | Fault Detection | depending on the initial solutions and easily trapping into the local optima | having low computational complexity |
| **MOABC (Hancer, Xue et al. 2015)** | MOP | Wrapper | ABC | Numeric | Inefficient for high-dimensional datasets due to high computational complexity. | Low probability of trapping into the local optimum, High classification accuracy |
| **TMABC-FS(Zhang, Cheng et al. 2019)** | MOP | Wrapper | ABC | Numeric | | |
| **MOPSOFS(Xue, Zhang et al. 2013)** | MOP | Wrapper | PSO | Numeric | | |
| **MOACO (Ke, Feng et al. 2010)** | MOP | Wrapper | ACO | Numeric | | |
| **MOWGA (Vignolo, Milone et al. 2013)** | MOP | Wrapper | GA | Face Recognition | | |
| **MODEFS (Mlakar, Fister et al. 2017)** | MOP | Wrapper | DE | Face Recognition | | |
| **MONSGA (González, Ortega et al. 2019)** | MOP | Wrapper | NSGA-II | Motor imagery | | |
| **Nested GA (Sayed, Nassef et al. 2019)** | MOP | Wrapper | NSGA-II | Microarray | | |

## 3. Swarm intelligence-based feature selection

Feature selection methods based on how to evaluate the features are classified into two categories: feature ranking and subset selection methods. Feature ranking methods, based on a specified criterion, assign a score to each feature. Then the features that did not get enough scores are removed. But in subset selection methods, the space of possible subsets is searched to find the optimal subset. If the number of initial features is $n$, the search space for the optimal subset contains all the subsets of the features, which is equal to $2^n$ different states. In other words, in the property ranking methods, the value of each property is evaluated independently, and the relationship between features is not considered. In these methods, it is assumed that the features are independent of each other and that a possible dependence between the features is not considered. Although this simplistic assumption reduces the computational complexity of the feature selection method, in many cases, it may reduce the performance of the feature selection method.

Feature subset selection is an NP-Hard problem. In the simplest way, the best subset can be found by evaluating all possible subsets with an exhaustive search strategy. Although this method guarantees an optimal feature subset, finding the optimal solution is very time consuming and even impractical even for medium-sized datasets. Since it is very costly to evaluate all possible subsets, a feature subset must be searched, that, it is acceptable both in terms of computational complexity and in terms of appropriateness. (Li, Chen et al. 2020, Santucci, Baioletti et al. 2020). One approach to solve complex optimization and NP-Hard problems is meta-heuristics algorithms. Meta-heuristic algorithms are approximate approaches that can find satisfactory solutions over acceptable time instead of finding the optimal solution (Zhang, Lee et al. 2015). These algorithms are one of the categories of approximate optimization algorithms that have s strategies to escape from local optima and can be used in a wide range of optimization problems.

Many feature selection methods use meta-heuristics to avoid increasing computational complexity in the high dimensional dataset. These algorithms use primitive mechanisms and operations to solve an optimization problem and search for the optimal solution over a number of iterations (Barak, Dahooie et al. 2015). These algorithms often start with a population containing random solutions and try to improve the optimality of these solutions during each iteration step. At the beginning of most of the meta-heuristic algorithms, a number of initial solutions are randomly generated, and then a fitness function is utilized to calculate the optimality of the individual solutions of the generated population. If none of the termination criteria are met, production new generation will begin. This cycle is repeated until one of the termination criteria is met (Hu, Zheng et al. 2020, Wang, Pan et al. 2020).

Meta-heuristic approaches can be classified into two categories: The Evolutionary Algorithms (EA) and Swarm Intelligence (SI) (Zhang, Lee et al. 2015). An EA uses mechanisms inspired by biological evolution, such as reproduction, mutation, recombination, and selection. Candidate solutions to the optimization problem play the role of individuals in a population, and the fitness function determines the quality of this solutions. After repetitions of the evolutionary algorithm, the initial population evolves and moves toward global optimization(Gong, Xu et al. 2020). On the other, SI algorithms usually consist of a simple population of artificial agents locally with the environment. This concept is usually inspired by nature, and each agent performs an easy job, but local interactions and partly random interactions between these agents lead to the emergence of "intelligent" global behavior, which is unknown to individual agents(Yong, Dun-wei et al. 2016).

In the remainder of this section, SI-based feature selection methods, such as Particle Swarm Optimization (PSO), Ant Colony Optimization (ACO), Artificial Bee Colony optimization (ABC), Differential Evolution (DE), Gravitational Search Algorithm (GSA), Firefly Algorithm (FA), Bat Algorithm (BA), Cuckoo Optimization Algorithm (COA), Gray Wolf Optimization (GWO), Whale Optimization Algorithm (WOA) and Salp Swarm Algorithm (SSA) are reviewed and outlined in Table 2.

## 3.1. PSO-based methods

Particle Swarm Optimization is an efficient swarm intelligence-based evolutionary algorithm, introduced by Kennedy and Eberhart in 1995 (J. Kennedy and Eberhart 1995). The PSO algorithm, inspired by the social behavior of birds and fish, has recently been utilized in many studies to solve the feature selection problem. Under et al. (Unler, Murat et al. 2011) proposed a feature selection method, which combined the univariate filter model within the PSO-based wrapper model. In (Inbarani, Azar et al. 2014), another hybrid model is proposed to feature selection in the medical application. In this method, hybrid feature selection based on PSO and rough sets theory are utilized to improve disease diagnosis in the medical dataset. Moreover, Huang and Dun (Huang and Dun 2008) propose a hybrid model for feature selection and parameter optimization integration, the PSO algorithm, and support vector machines (SVM) classifier. To overcome the high computational complexity in the high dimensional dataset, in this paper the combined PSO–SVM is implemented with a distributed parallel architecture. Furthermore, in (Xue, Zhang et al. 2014) propose new different initialization strategies and new best particle updating mechanisms are proposed to improve feature selection accuracy in the classification task. The purpose of this proposed method is to reduce the number of selected features and thus reduce the computational complexity of the data classification process. In (Banka and Dara 2015), a Hamming distance-based binary particle swarm optimization (HDBPSO) algorithm is proposed to reduce data dimensions. In this method, a hamming distance is utilized to update the velocity of particles in a binary PSO search strategy to select the essential features. In (Yong, Dun-wei et al. 2016), and improved multi-objective PSO algorithm is proposed for unreliable data classification. In this paper, two new operators of the reinforced memory and the hybrid mutation are introduced to improve the search ability of the PSO algorithm. In (Moradi and Gholampour 2016), and efficient PSO-based feature selection method is proposed by the integration of filter and wrapper approaches. The proposed method, called HPSO-LS, introduced a new local search to select the subset of non-redundant and relevant features. In (Zhang, Gong et al. 2017) a multi-Objective particle swarm optimization approach is developed for cost-based feature selection in classification. The main goal of this method is to produce a Pareto front of nondominated solutions, that is, feature set, to meet different prerequisites of decision-makers in real-world applications. In (Jain, Jain et al. 2018), integration of correlation feature selection with modified binary PSO algorithm is used for gene selection and cancer classification. This method selects a high relevant feature subset by eliminating the irrelevant and redundant features. In (Zhang, Ding et al. 2018), a combination method of improved mRMR and Shuffled

Frog Leaping Algorithm is developed to improve the acoustic defect detection accuracy. In this method, to reduce the dimensions of the original features, a single-objective function is defined using the MRMR criterion and then this function is optimized using the PSO algorithm. Finally, after selecting the final feature subset, the final acoustic defect detection is made using the neural network classifier. The authors of (Qasim and Algamal 2018) proposed a PSO-based regression model for dimensionality reduction in medical dataset. This method makes use of the advantages of both PSO and the logistic regression with Bayesian information criterion as a fitness function. In (Prasad, Biswas et al. 2018), a recursive PSO scheme for feature selection in DNA microarray datasets is proposed. In this method, various filter-based ranking methods with the proposed recursive PSO based wrapper approach are integrated. Pashaei et al. (Pashaei, Pashaei et al. 2019) proposed the hybrid feature selection method in cancer classification using binary black hole algorithm and modified binary particle swarm optimization. In this method, the binary black hole algorithm is embedded in the PSO algorithm to make this more effective and to facilitate the exploration and exploitation of the PSO to improve the performance further. Moreover, in (Gunasundari, Janakiraman et al. 2018), a multi swarm sophisticated binary particle swarm optimization algorithm using a Win-Win approach for feature selection of liver and kidney cancer data is proposed. Moreover, In (Yan, Liang et al. 2019), a hybrid feature selection method based on PSO was proposed to improve the accuracy of Laser-induced breakdown spectroscopy analysis. This method integrates the advantages of the filter and wrapper model. The filter model first eliminated the uncorrelated and redundant features, and then the selected features were further refined by a more accurate wrapper method PSO. In (Xue, Tang et al. 2020) a PSO-based feature selection with multiple classifiers is proposed for improve for increasing the classification accuracy and reducing computational complexity. In this paper, a new Self-Adaptive Parameter and Strategy are used to deal with the issue of feature selection in high-dimensional dataset. The reported results showed that the use of these mechanisms greatly increased the search ability of particle optimization algorithms for high-dimensional datasets.

## 3.2. ACO-based methods

In the early 1990s, an algorithm called the Ant System (AS) was proposed as a new heuristic for solving difficult optimization problems by Dorigo and his colleagues and was first applied to the TSP problem. The Ant Colony Optimization Algorithm (ACO) was proposed by Dorrigo and his colleagues as a multi-agent to solve the optimization problems (M. Dorigo  and Caro 1999). This algorithm is inspired by the behavior of ants that are able to find the shortest path between the nest and the food source and also adapt to environmental changes. Moreover, ACO has been successfully applied in several studies to feature selection. Kabir et al. (Monirul Kabir, Shahjahan b et al. 2012) propose a new ACO-based feature selection method by the integration of neural network and information gain to select a subset of salient features of reduced size. In (Ying Li, Gang Wang et al. 2013), a framework based on the ACO algorithm for gene selection in the DNA microarray dataset, which consists of two stages of pivotal genes choosing by modified ant system and the most important genes selecting by the modified ant colony system. Moreover, Chen et al. (Chen, Chen et al. 2013) propose a feature selection method based on the ACO algorithm. In this paper, different from previous ACO-based methods, the feature space represented as a directed graph. In (Forsati, Moayedikia et al. 2014), a novel variant of ACO was introduced to select the most relevant and non-redundant features. This algorithm is combined with a local search strategy to overcome the entrapment in local optima by searching the neighborhood of the globally optimal solution. On the other hand, in (Ke, Feng et al. 2008), an ACO-based filter model is proposed for feature selection in rough set theory. Tabakhi et al. (Tabakhi, Moradi et al. 2014) propose an unsupervised feature selection method based on ant colony optimization. In this method, the final features are selected considering the similarity between the features and without using a learning algorithm. Moreover, in (Moradi and Rostami 2015), a feature selection method using the ant colony algorithm and social network analysis technique is proposed. In this method, that a clustered graph is used to represent the feature selection problem, no learning algorithm is used to evaluate the generated feature subset, and the appropriateness of each solution is evaluated using a filter criterion. Zamani Dadaneh et al. (Dadaneh, Markid et al. 2016) proposed unsupervised probabilistic feature selection using ACO algorithm. The algorithm looks for the optimal feature subset in an iterative process by utilizing the similarity between the features. In (Liu, Wang et al. 2019) combination of feature selection and ant colony optimization is proposed for improve the classification accuracy of imbalanced data. In this method, instead of using a single-objective fitness function, a multi-objective ant colony optimization algorithm is used to improve the performance feature selection. The reported results showed acceptable performance of the proposed method in classifying imbalanced and high-dimensional datasets.

### 3.3. ABC-based methods

The Artificial Bee Colony algorithm (ABC) is an optimization algorithm based on swarm intelligence and intelligent behavior of the bee population that simulates the food search behavior of bee groups. In the early version of this algorithm, it performs a kind of local search that is combined with a random search and can be used for hybrid optimization or functional optimization. This SI-based algorithm has been utilized in many studies to search for the optimal feature subset. In (Mauricio Schiezaro and Pedrini 2013), an ABC-based data feature selection is proposed to improve the classification accuracy. In this method, when the feature selection problem is represented as a binary vector, the classification accuracy in the classifier is used as a fitness function. In (Hancer, Xue et al. 2015), a modified artificial bee colony algorithm for feature selection problems is proposed. In this method, similarity search mechanisms with existing binary ABC variant is integrated for improving the feature selection accuracy. In (Shunmugapriya and Kanmani 2017), Artificial Bee Colony with Ant Colony Optimization is integrated to optimize feature subset selection. This hybrid algorithm makes use of the advantages of both ACO and the ABC algorithm, and the results show the promising behavior of the proposed algorithm. Cancer et al. (Hancer, Xue et al. 2018) proposed a feature selection method by the combination of the ABC algorithm and the pareto-optimal front surface. The authors of this paper used a multi-objective fitness function and genetic operators to improve the accuracy of the feature selection method and the convergence of the ABC algorithm. The authors of (Arslan and Ozturk 2019) proposed a Multi Hive Artificial Bee Colony Programming for high dimensional feature selection. In this method, the ability of automatic programming methods to select truly relevant features is investigated in training and test data sets. In (Zhang, Cheng et al. 2019), a feature selection method using a two-archive multi-objective artificial bee colony algorithm is proposed. In this method, two new operators and are utilized to enhance its search capability and convergence. In (Wang, Zhang et al. 2020), an ABC-based feature selection is proposed by integrating of multi-objective optimization algorithm with a sample reduction strategy. This proposed method has both increased classification accuracy and reduced computational complexity.

### 3.4. DE-based methods

The Differential Evolution (DE) algorithm is an evolutionary algorithm based on swarm intelligence that has been proposed to solve various optimization problems. This optimization algorithm is proposed to overcome the main flaw of the genetic algorithm, namely the lack of local search in this algorithm and the main difference between it and the genetic algorithms is in the genetic selection operators. DE algorithm is utilized in different applications of pattern recognition and feature selection. For example, Al-Ani et al. (Al-Ani, Alsukker et al. 2013) proposed a wrapper-based feature subset selection using differential evolution and a wheel based search strategy. Hancer et al. (Hancer, Xue et al. 2018) developed a multi-objective differential evolution-based method to select the most relevant feature set and improving classification accuracy by eliminating irrelevant and redundant features. In (Zhang, Gong et al. 2020), a multi-objective feature selection approach called the Binary Differential Evolution with self-learning (MOFS-BDE) is proposed. In this method, a new mutation operator is introduced to escape the local optimal based. Furthermore, in (Hancer 2020), a new multi-objective differential evolution approach feature selection is proposed to feature selection and improve the performance of the clustering algorithm simultaneously.

### 3.5. GSA-based methods

Moreover, some Swarm Intelligence algorithms have been inspired by physical laws. Gravitational Search Algorithm (GSA) is (Rashedi, Nezamabadi-pour et al. 2009) a physics-based algorithm inspired by Newton's law of universal gravitation. GSA, a popular swarm intelligence technique, has been widely employed in data management, and recently, many feature selection methods have been proposed using this optimization algorithm. Han et al. (Han, Chang et al. 2014) developed a feature selection method by the integration of the GSA algorithm with a linear chaotic map for improving classification accuracy. In (Xiang, Han et al. 2015), a hybrid system for feature selection based on an improved gravitational search algorithm and k-NN method is proposed. In this method, a piecewise linear chaotic map for exploration, and sequential quadratic programming for exploitation are utilized. In (Taradeh, Mafarja et al. 2019), and efficient hybrid feature selection method is introduced to utilized the advantages of the Swarm Intelligence algorithms and the genetic algorithm for improving the performance of the gravitational search algorithm.

### 3.6. FA-based methods

The Firefly Algorithm (FA) was introduced by Xin-She Yang, in 2010 (Yang 2010), that, the main idea was inspired by the optical connection between fireflies. The Firefly Algorithm is a sensible example of swarm intelligence in which low-performance agents can work together to achieve great results with high performance. In (Zhang, Song et al. 2017) a novel FA-based feature selection method, called return-cost-based binary FFA is developed. The authors of this paper provide a variety of strategies to prevent premature convergence of the FA algorithm and thus improve the accuracy of feature selection. In (Zhang, Mistry et al. 2018), a feature selection method is developed using the firefly optimization algorithm to increase the accuracy of classification and regression models. In this method, to improve convergence and prevent trapping in local optimization, some variation is added to standard FA, which improves the accuracy of final feature selection. Moreover, in (Larabi Marie-Sainte and Alalyani 2020) FA-based feature selection method is proposed. In this method, after selecting the final features, these features are utilized to classify Arabic texts using an SVM classifier. Furthermore, in (Selvakumar and Muneeswaran 2019), another FA-based feature selection method is presented for network intrusion detection. In this method, a combination of filter-based feature selection method (i.e., Mutual Information) and wrapper-based feature selection method (i.e., C4.5 and Bayesian network) has been utilized to select the final features.

### 3.7. BA-based methods

The Bat Algorithm (BA) (Yang 2010) is an algorithm inspired by the collective behavior of bats in the natural environment, introduced by Yang in 2010. The bat algorithm is a kind of swarm intelligence-based algorithm that is inspired by the echolocation behavior of bats. Bats find the exact path and location of their prey by sending sound waves and receiving reflections. When the sound waves return to the transmitter of the bat waves, the bird can draw an audio image of the obstacles in front of its surroundings and see the surroundings well. In (Tawhid and Dsouza 2018), a hybrid variant of bat algorithm and improved PSO algorithm is to improve the feature selection performance. In this proposed method, the PSO algorithm is used to improve the convergence power of the hybrid algorithm. Moreover, in (Liu, Yan et al. 2020), a binary BA-based feature selection method for image steganalysis is proposed. This method selects the most relevant feature from raw features extracted to improve the final detection accuracy. Furthermore, Azmi Al-Betara et al. (Al-Betar, Alomari et al. 2020) used the bat algorithm to search optimal features subset to increase the accuracy of cancer classification. In this method, a combination of filter and wrapper approach was used to improve the performance of feature selection. In this method, robust mRMR as a filter to select the most relevant features and an improved BA algorithm as a search strategy in the wrapper approach is presented to select the final feature subset.

## 3.8. COA-based methods

The Cuckoo Optimization Algorithm (COA) is another algorithm based on swarm intelligence, inspired by the special lifestyle of a bird called the cuckoo (Rajabioun 2011). The specific habitude of laying eggs and the reproduction of this bird has been the basis for the formation of this optimization algorithm. Like other evolutionary algorithms, the cuckoo optimization algorithm begins with an initial population of cuckoos. This early population of cuckoos has a number of eggs that are placed in the nest of the host bird. Some eggs, which are more similar to host bird eggs, have a better chance of growing into adult birds. The other eggs are identified and destroyed by the host butterfly. Grown eggs show that the nest is a better place in the search space, and the usefulness of that area is higher. The goal of the cuckoo optimization algorithm, which is the optimization function, is to find the place where most eggs have a more chance to survive. In (Elyasigomari, Lee et al. 2017), a COA-based feature selection method is developed to improve the cancer classification accuracy. In this method, first, the irrelevant features are removed using a simple and fast filter-based feature selection. Then, from these relevant features, the final features are selected by integration wrapper-based feature selection and COA algorithm. In (Jayaraman and Sultana 2019), a combination method of cuckoo search algorithm and neural network is developed for feature selection. In this method, after selecting a feature subset of non-redundant and relevant features, the final chosen features are sent to the classifier, and heart disease is classified. Moreover, in (Prabukumar, Agilandeeswari et al. 2019), another cuckoo search-based feature selection method is proposed for improving the disease diagnosis accuracy. In this method, the process of feature subset selection is optimized using the cuckoo search optimization, and then these selected features are sent to the SVM classifier to identify lung cancer.

## 3.9. GWO-based methods

Gray Wolf Optimization (GWO) is a new meta-heuristic algorithm inspired by grey wolves (Mirjalili, Mirjalili et al. 2014). This algorithm is one of the latest bio-inspired techniques, which simulate the hunting process of a pack of gray wolves in nature. The GWO algorithm mimics the leadership hierarchy and hunting mechanism of grey wolves in nature. Recently, some GWO-based method has been used as a tool for feature selection in data mining. In (Emary, Yamany et al. 2015), a Multi-Objective GWO was employed to search the search most relevant and non-redundant features. This proposed hybrid method uses the low computational complexity of the filter model to improve the performance of the wrapper model. In (Emary, Zawbaa et al. 2016), a binary version of the GWO is proposed to select the optimal feature subset for classification tasks. In the wrapper-based feature selection method, classification accuracy, and the number of selected features are utilized for the fitness function. In (Tu, Chen et al. 2019), to improve the previous GWO-based method, a multi-strategy ensemble GWO is proposed for feature selection. Moreover, Abdel-Basset et al. (Abdel-Basset, El-Shahat et al. 2020) proposed a Grey Wolf Optimizer wrapper-based feature selection method integrated with a Mutation operator for data classification. The mutation operator presented in this paper tries to reduce the features that increase the redundancy between the selected features and add the features that increase the classification accuracy.

## 3.10. WOA-based methods

Whale Optimization Algorithm (WOA) is another new swarm-based optimization algorithm inspired by the hunting behavior of humpback whales (Mirjalili and Lewis 2016). This algorithm included three operators to simulate the search for prey, encircling prey, and bubble-net foraging behavior of humpback whales. Recently, the Whale Optimization Algorithm is successfully utilized in many different optimization problems. In (Mafarja and Mirjalili 2017), a hybrid WOA with a simulated annealing method is proposed for feature selection. The goal of using simulated annealing in this hybrid method is to enhance the exploitation by searching the most promising regions located by the WOA algorithm. In (Mafarja and Mirjalili 2018), a wrapper feature selection approach is proposed based on WOA. In this work, tournament and roulette wheel selection strategy and also crossover and mutation operators are utilized to enhance the exploration and exploitation of the search process of swarm intelligence algorithms. Nematzadeh et al. (Nematzadeh, Enayatifar et al. 2019) proposed a frequency-based filter feature selection method using whale algorithm on high dimensional medical datasets. In this method, a filter criterion is used to discard the irrelevant features using the WOA. Then, the reminder features are ranked based on another filtering method, namely, Mutual Congestion.

## 3.11. SSA-based methods

Salp Swarm Algorithm (SSA) (Mirjalili, Gandomi et al. 2017) another bio-inspired algorithm based on swarm intelligence is proposed for solving optimization problems. This algorithm is inspired by the swarming behavior of salps when moving and forage in the oceans. In (Faris, Mafarja et al. 2018) an efficient binary Salp Swarm Algorithm with crossover scheme is proposed to improve the accuracy of feature selection. Moreover, in (Ibrahim, Ewees et al. 2019) a hybrid optimization method for the feature selection problem is proposed by integration of the slap swarm algorithm with the particle swarm optimization. this combination between both swarm optimization algorithms improved the efficacy of the exploration and the exploitation steps. In (Tubishat, Idris et al. 2020), integration of Improved SSA and novel local search algorithm is developed for the feature selection problem. In this method, the accuracy of the classification accuracy of KNN classifier on the training data is used as a fitness function, and the final features are selected in a repetitive process. Also, in this method, to create the initial population, a method based on Opposition Based Learning is presented, which has caused diversity in the initial solutions. Furthermore, in (Al-Zoubi, Heidari et al. 2020) a Salp Swarm Algorithm-based feature weighting method is developed to predict the presence of liver disorder, heart, and Parkinson’s disease. Also, in (Hegazy, Makhlouf et al. 2020) solution accuracy, reliability and convergence speed of basic SSA is improved by adding a new control parameter and inertia weight. Then this improved algorithm is tested in feature selection problem. Moreover, in (Neggaz, Ewees et al. 2020), a new variant of SSA optimization algorithm is proposed for feature selection. In this method, an additional phase is utilized to overcome the problem of being stuck in local optima by encouraging the exploration of the search.

**Table 2**: Outlining the reviewed swarm intelligence based feature selection methods

| Authors | SI method | Number of Objectives | Type | Application |
|---|---|---|---|---|
| (Unler, Murat et al. 2011) | PSO | Single Objective | Hybrid | Numerical |
| (Inbarani, Azar et al. 2014) | PSO | Single Objective | Hybrid | Medical |
| (Huang and Dun 2008) | PSO | Single Objective | Wrapper | Numerical |
| (Xue, Zhang et al. 2014) | PSO | Single Objective | Wrapper | Numerical |
| (Banka and Dara 2015) | PSO | Multi Objective | Wrapper | Medical |
| (Yong, Dun-wei et al. 2016) | PSO | Multi Objective | Wrapper | Numerical |
| (Moradi and Gholampour 2016) | PSO | Single Objective | Wrapper | Numerical/Medical |
| (Zhang, Gong et al. 2017) | PSO | Single Objective | Wrapper | Numerical |
| (Jain, Jain et al. 2018) | PSO | Single Objective | Hybrid | DNA microarray |
| (Zhang, Ding et al. 2018) | PSO | Single Objective | Filter | Signal |
| (Qasim and Algamal 2018) | PSO | Single Objective | Wrapper | Medical |
| (Prasad, Biswas et al. 2018) | PSO | Single Objective | Wrapper | DNA microarray |
| (Pashaei, Pashaei et al. 2019) | PSO | Single Objective | Wrapper | Medical |
| (Gunasundari, Janakiraman et al. 2018) | PSO | Single Objective | Wrapper | Medical |
| (Yan, Liang et al. 2019) | PSO | Single Objective | Wrapper | spectroscopy |
| (Xue, Tang et al. 2020) | PSO | Single Objective | Wrapper | Numerical |
| (Monirul Kabir, Shahjahan b et al. 2012) | ACO | Single Objective | Hybrid | Numerical |
| (Ying Li, Gang Wang et al. 2013) | ACO | Single Objective | Wrapper | DNA microarray |
| (Chen, Chen et al. 2013) | ACO | Single Objective | Wrapper | Image |
| (Forsati, Moayedikia et al. 2014) | ACO | Single Objective | Wrapper | Numerical |
| (Ke, Feng et al. 2008) | ACO | Single Objective | Filter | Numerical |
| (Tabakhi, Moradi et al. 2014) | ACO | Single Objective | Filter | Numerical/Medical |
| (Moradi and Rostami 2015) | ACO | Single Objective | Filter | Numerical/Medical |
| (Dadaneh, Markid et al. 2016) | ACO | Single Objective | Filter | Numerical/Medical |
| (Liu, Wang et al. 2019) | ACO | Multi Objective | Wrapper | Numerical |
| (Mauricio Schiezaro and Pedrini 2013) | ABC | Single Objective | Wrapper | Numerical/Medical |
| (Hancer, Xue et al. 2015) | ABC | Single Objective | Wrapper | Numerical/Medical |
| (Shunmugapriya and Kanmani 2017) | ABC | Single Objective | Wrapper | Numerical/Medical |
| (Hancer, Xue et al. 2018) | ABC | Single Objective | Wrapper | Numerical |
| (Arslan and Ozturk 2019) | ABC | Single Objective | Wrapper | Numerical |
| (Zhang, Cheng et al. 2019) | ABC | Multi-Objective | Wrapper | Numerical |
| (Wang, Zhang et al. 2020) | ABC | Multi-Objective | Wrapper | Numerical |
| (Al-Ani, Alsukker et al. 2013) | DE | Single Objective | Wrapper | Numerical/Medical |
| (Hancer, Xue et al. 2018) | DE | Multi Objective | Filter | Numerical/Medical |
| (Zhang, Gong et al. 2020) | DE | Multi Objective | Wrapper | Numerical/Medical |
| (Hancer 2020) | DE | Single Objective | Wrapper | Numerical/Medical |
| (Han, Chang et al. 2014) | GSA | Single Objective | Wrapper | Numerical/Medical |
| (Xiang, Han et al. 2015) | GSA | Single Objective | Wrapper | Numerical/Medical |
| (Taradeh, Mafarja et al. 2019) | GSA | Single Objective | Wrapper | Numerical/Medical |
| (Zhang, Song et al. 2017) | FA | Multi-Objective | Wrapper | Numerical |
| (Zhang, Mistry et al. 2018) | FA | Single Objective | Wrapper | Facial expression |
| (Larabi Marie-Sainte and Alalyani 2020) | FA | Single Objective | Wrapper | Text |
| (Selvakumar and Muneeswaran 2019) | FA | Single Objective | Hybrid | Network Intrusion |
| (Tawhid and Dsouza 2018) | BA | Single Objective | Wrapper | Numerical/Medical |
| (Liu, Yan et al. 2020) | BA | Single Objective | Wrapper | Image Steganalysis |
| (Al-Betar, Alomari et al. 2020) | BA | Single Objective | Hybrid | Medical |
| (Elyasigomari, Lee et al. 2017) | COA | Single Objective | Wrapper | Numerical/Medical |
| (Jayaraman and Sultana 2019) | COA | Single Objective | Wrapper | Medical |
| (Prabukumar, Agilandeeswari et al. 2019) | COA | Single Objective | Wrapper | Gene expression |
| (Emary, Yamany et al. 2015) | GWO | Multi-Objective | Hybrid | Numerical/Medical |
| (Emary, Zawbaa et al. 2016) | GWO | Single Objective | Wrapper | Numerical/Medical |
| (Tu, Chen et al. 2019) | GWO | Single Objective | Wrapper | Numerical/Medical |
| (Abdel-Basset, El-Shahat et al. 2020) | GWO | Single Objective | Wrapper | Numerical/Medical |
| (Mafarja and Mirjalili 2017) | WOA | Single Objective | Wrapper | Numerical/Medical |
| (Mafarja and Mirjalili 2018) | WOA | Single Objective | Wrapper | Numerical/Medical |
| (Nematzadeh, Enayatifar et al. 2019) | WOA | Single Objective | Filter | Medical |
| (Faris, Mafarja et al. 2018) | SSA | Single Objective | Wrapper | Numerical/Medical |
| (Ibrahim, Ewees et al. 2019) | SSA | Single Objective | Wrapper | Numerical |
| (Tubishat, Idris et al. 2020) | SSA | Single Objective | Wrapper | Numerical/Medical |
| (Al-Zoubi, Heidari et al. 2020) | SSA | Single Objective | Wrapper | Medical |
| (Hegazy, Makhlouf et al. 2020) | SSA | Multi-Objective | Wrapper | Numerical/Medical |
| (Neggaz, Ewees et al. 2020) | SSA | Single Objective | Wrapper | Numerical/Medical |

## 4. Experimental results

In this section, the performances of the different swarm intelligence-based feature selection methods are evaluated. The results are shown in terms of the number of selected features and the classification accuracy (ACC). The classification accuracy calculated as follows:

$$ACC = \frac{TP + TN}{TP + TN + FP + FN} \qquad (1)$$

Where TP, TN, FP, and FN stand for the number of true positives, true negatives, false positives, and false negatives, respectively.

In each experiment, each feature selection method is run ten times, and the mean and standard deviation (square root of its variance) of ten different runs is used to compare different methods. Moreover, in each run, each dataset is normalized, and it is randomly divided into a training set (66 % of the dataset) and a test set (34 % of the dataset). The training set is utilized for the feature selection process, while the test set is applied for evaluating the proposed feature selection method. To fulfill fair experiments, all evaluated methods are carrying out on the same train/test dataset. Due to the randomness of the train and test set, both the average and the standard deviation of the result are reported. In the remainder of this section, used datasets, utilized classifiers, evaluated methods, results, and discussion are explained in the following subsections.

## 4.1. Datasets

In this study, several datasets, with different specifications, were utilized to evaluate different SI-based feature selection methods and compare their performance with each other. These datasets include SpamBase, Sonar, Arrhythmia, Madelon, Isolet, and Colon taken from the UCI repository (Asuncion and Newman 2007) and have been extensively used in the literature. The basic characteristics of these datasets are summarized in Table 3. These datasets have been chosen considering diverse characteristics such as the number of features and number of different classes. For example, Colon is a very high dimensional dataset with a small sample size, while SpamBase is the example of a low dimensional, with a large sample size dataset. Again Isolet is a multi-class dataset that has 26 different kinds of classes.

In some of these datasets, different features have different values. In this situation, features with a larger value range may dominate the features with a smaller value range, and maybe more likely to be selected. To overcome this challenge, before the feature selection process, all different datasets are normalized using the Max-Min normalization. Using this normalization method, the range of values of all used datasets is changed to [0 1] range.

Moreover, in some of the used datasets, there are several missing values. To overcome this difficulty, the miss values in these features are inserted by averaging the available data corresponding to the available features.

**Table 3:** Characteristics of the used medical datasets

| Dataset | Features | Classes | Patterns |
|---|---|---|---|
| **SpamBase** | 57 | 2 | 4601 |
| **Sonar** | 60 | 2 | 208 |
| **Arrhythmia** | 279 | 16 | 351 |
| **Madelon** | 500 | 2 | 4400 |
| **Isolet** | 617 | 26 | 1559 |
| **Colon** | 2000 | 2 | 62 |

## 4.2. The utilized Classifier

To evaluate the generalizability of the proposed methods in different classifiers, in these experiments, three classifiers, including Support Vector Machine (SVM), Naïve Bayes (NB), and AdaBoost (AB) are utilized.

Support vector machine SVM is one of the supervised learning algorithms that was proposed by Vapnik. The goal of SVM is to maximize the margin between data samples, and in recent years it has shown good performance for classification and regression problems. Naïve Bayes (NB) is a group of simple probabilistic classification algorithms based on the probability that classifies data by assuming the independence of random variables and using the base theorem. AdaBoost (AB), short for "Adaptive Boosting", which is a machine learning meta-algorithm formulated by Yoav Freund and Robert Schapire. An AdaBoost classifier is a meta-estimator that begins by fitting a classifier and then fits additional copies of it on the same dataset, and then the weights of incorrectly classified instances are adjusted such that subsequent classifiers focus more on severe cases.
The experimental workbench is Weka (Waikato environment for knowledge analysis) (Hall, Frank et al.), which is a collection of data mining methods. In this work, SMO, Naïve Bayes, and AdaBoostM1 as the WEKA implementation of SVM, NB, and AB were used.

## 4.3. The evaluated methods

In these experiments, to compare the performance of different methods of selecting a feature based on, from each SI-based algorithm, one feature selection method was chosen and evaluated in the experimental result. For a fair evaluation, all of the methods examined in this section were selected from among wrapper-based methods. Due to the inherent differences between filter-based and wrapper-based feature selection methods, which usually filter-based model have lower computational complexity and wrapper-based models have higher accuracy, it is not possible to compare these methods. For a fairer comparison, the performed experiments in this section are divided into two separate parts, in the first part the wrapper-based methods and in the second part the filter-based methods were compared with each other.

These wrapper-based methods include PSO-based (Xue, Tang et al. 2020), ACO-based (Liu, Wang et al. 2019), ABC-based (Wang, Zhang et al. 2020), DE-based (Hancer 2020), GSA-based (Taradeh, Mafarja et al. 2019), FA-based (Selvakumar and Muneeswaran 2019), BA-based (Tawhid and Dsouza 2018), COA-based (Elyasigomari, Lee et al. 2017), GWO-based (Abdel-Basset, El-Shahat et al. 2020), WOA-based (Mafarja and Mirjalili 2018) and SSA-based (Neggaz, Ewees et al. 2020). Furthermore, filter-based methods include PSO-based (Zhang, Ding et al. 2018), ACO-based (Moradi and Rostami 2015), DE-based. (Hancer, Xue et al. 2018) and WOA-based (Nematzadeh, Enayatifar et al. 2019).

Moreover, all these methods are implemented using C# on an Intel Core-i7 CPU with 8GB of RAM.

## 4.4. Results

In these experiments, the classification accuracy and feature subset size are used as the performance evaluation criteria. In the experiments, first, the performances of different wrapper SI-based feature selection methods are compared over different classifiers. Tables 4 summarize the average classification accuracy (in %) over ten independent runs of the different SI-based wrapper feature selection methods using SVM, NB, and AB classifiers. Each entry of the Tables 4 denotes the mean value as well as standard deviation (shown in parenthesis) of ten independent runs. The best result is indicated in bold face and underlined, and the second-best is in bold face. Table 4 reveals that, in most cases, the PSO-based method performs better than the other SI-based feature selection method. For example, in SpamBase dataset on the SVM classifier, PSO-based method obtained a 92.84 % classification accuracy. In contrast, for ACO, ABC, DE, GSA, FA, BA, COA, GWO, WOA, and SSA-based method, these values were reported.91.87, 90.32, 88.78, 88.09, 87.91, 90. 83, 91.16, 91.73, 88.64 and 90.19, correspondingly.

**Table 4:** Average classification accuracy rate and as standard deviation (shown in parenthesis) over ten runs of the wrapper-based feature selection methods using SVM, Naive Bayes, and AdaBoost classifier. The best result is indicated in bold face and underlined, and the second-best is in bold face.

| Dataset | Method | Classifier |
| --- | --- | --- |

| | | SVM | Naive Bayes | AdaBoost |
|---|---|---|---|---|
| **SpamBase** | PSO | **<u>92.84 (2.83)</u>** | **<u>92.31 (1.56)</u>** | **<u>92.75 (2.35)</u>** |
| | ACO | **91.87 (1.83)** | 89.76 (2.82) | 90.66 (2.91) |
| | ABC | 90.32 (2.09) | 89.91 (1.93) | **91.91 (3.13)** |
| | DE | 88.78 (2.54) | 88.32 (2.19) | 89.94 (2.87) |
| | GSA | 88.09 (1.96) | 87.98 (2.02) | 88.27 (2.62) |
| | FA | 87.91 (2.17) | **92.04 (1.82)** | 90.45 (1.93) |
| | BA | 90. 83 (2.94) | 89.73 (2.37) | 90.65 (1.76) |
| | COA | 91.16 (1.65) | 90.83 (2.74) | 90.77 (1.84) |
| | GWO | 91.73 (1.87) | 91.85 (3.09) | 91.84 (2.90) |
| | WOA | 88.64 (2.37) | 88.93 (2.18) | 90.38 (1.45) |
| | SSA | 90.19 (2.26) | 90.06 (1.78) | 90.19 (2.89) |
| **Sonar** | PSO | 88.14 (2.47) | **87.91 (2.29)** | **<u>86.93 (3.11)</u>** |
| | ACO | **88.78 (2.87)** | 87.32 (3.42) | 85.81 (2.63) |
| | ABC | **<u>87.93 (1.82)</u>** | 87.12 (2.81) | **86.78 (1.15)** |
| | DE | 85.15 (2.78) | 85.19 (2.22) | 84.93 (2.81) |
| | GSA | 84.71 (1.69) | 85.11 (1.34) | 85.62 (1.83) |
| | FA | 84.92 (2.38) | 84.19 (2.41) | 86.62 (3.19) |
| | BA | 84.03 (1.83) | 84.66 (3.83) | 85.92 (2.58) |
| | COA | 84.73 (2.48) | **<u>89.84 (2.14)</u>** | 85.28 (1.94) |
| | GWO | 85.42 (1.19) | 85.31 (2.27) | 85.73 (2.08) |
| | WOA | 86.11 (2.78) | 85.65 (2.48) | 85.98 (2.69) |
| | SSA | 87.04 (3.02) | 87.26 (2.19) | 86.12 (2.73) |
| **Arrhythmia** | PSO | **<u>86.41 (2.71)</u>** | 86.05 (2.61) | **<u>85.94 (2.37)</u>** |
| | ACO | 84.92 (2.38) | **86.23 (2.75)** | **85.71 (2.83)** |
| | ABC | **85.14 (2.83)** | 85.72 (1.85) | 84.31 (1.86) |
| | DE | 82.28 (3.07) | 84.78 (2.39) | 83.81 (2.35) |
| | GSA | 81.79 (2.47) | 84.18 (2.61) | 84.21 (2.66) |
| | FA | 82.29 (1.74) | 83.28 (1.95) | 82.18 (1.97) |
| | BA | 83.36 (2.85) | 84.73 (2.91) | 83.46 (2.74) |
| | COA | 83.91 (2.79) | **<u>88.91 (2.63)</u>** | 82.31 (2.81) |
| | GWO | 82.05 (1.96) | 85.34 (1.79) | 84.92 (1.70) |
| | WOA | 83.38 (1.45) | 83.61 (2.94) | 83.38 (2.44) |
| | SSA | 81.84 (2.93) | 82.29 (2.09) | 82.86 (2.39) |
| **Madelon** | PSO | **86.67 (3.17)** | 86.63 (2.47) | **<u>86.16 (1.74)</u>** |
| | ACO | 86.19 (2.28) | 85.95 (1.38) | 86.03 (2.18) |
| | ABC | **<u>87.54 (2.73)</u>** | **<u>87.18 (1.81)</u>** | **86.08 (2.91)** |
| | DE | 85.14 (2.16) | 85.05 (2.81) | 85.92 (3.15) |
| | GSA | 84.38 (3.19) | 84.88 (1.92) | 84.67 (2.72) |
| | FA | 85.08 (2.82) | **86.95 (1.19)** | 85.81 (2.38) |
| | BA | 84.93 (1.84) | 85.07 (2.39) | 84.91 (1.96) |
| | COA | 85.26 (2.62) | 85.19 (1.83) | 85.83 (2.67) |
| | GWO | 84.27 (2.81) | 85.19 (1.92) | 85.37 (2.49) |
| | WOA | 84.93 (2.37) | 84.64 (2.37) | 85.04 (2.38) |
| | SSA | 85.09 (1.98) | 84.94 (1.64) | 84.77 (1.63) |
| **Isolet** | PSO | **<u>85.61 (1.33)</u>** | **<u>85.37 (1.92)</u>** | **<u>85.41 (2.31)</u>** |
| | ACO | **85.21 (1.91)** | **85.18 (1.82)** | **85.31 (2.39)** |
| | ABC | 84.32 (2.93) | 84.91 (2.36) | 84.51 (1.77) |
| | DE | 83.81 (2.92) | 83.04 (1.73) | 83.71 (2.37) |
| | GSA | 84.04 (1.86) | 84.23 (2.28) | 83.81 (1.73) |
| | FA | 84.78 (2.33) | 84.01 (1.82) | 84.66 (2.53) |
| | BA | 83.91 (1.87) | 83.19 (1.69) | 83.81 (1.48) |
| | COA | 85.01 (1.82) | 84.81 (2.83) | 85.01 (2.91) |
| | GWO | 84.32 (2.76) | 84.48 (1.91) | 84.39 (2.71) |
| | WOA | 84.51 (1.64) | 84.01 (2.19) | 83.92 (3.38) |
| | SSA | 85.19 (2.81) | 85.09 (2.31) | 84.93 (2.61) |
| **Colon** | PSO | **<u>96.41 (2.81)</u>** | **<u>96.19 (2.31)</u>** | **96.33 (2.39)** |
| | ACO | 96.11 (1.72) | 95.72 (1.81) | 96.15 (1.81) |
| | ABC | 93.71 (2.82) | 92.81 (3.92) | 92.81 (2.62) |
| | DE | 92.29 (2.38) | 91.18 (2,73) | 90.91 (1.88) |
| | GSA | 93.72 (1.84) | 92.19 (1.92) | 92.01 (1.03) |
| | FA | 90.37 (1.39) | 90.09 (2.71) | 91.26 (1.81) |
| | BA | 92.28 (1.67) | 91.26 (1.77) | 91.25 (2.70) |
| | COA | **96.57 (3.19)** | **95.88 (2.98)** | **<u>96.42 (1.78)</u>** |
| | GWO | 93.33 (2.82) | 92.85 (2.71) | 92.91 (3.11) |

| | | | |
|---|---|---|---|
| WOA | 91.19 (3.26) | 90.88 (1.61) | 91.27 (2.66) |
| SSA | 95.91 (2.18) | 94.81 (2.60) | 95.81 (2.36) |

Moreover, Figures 1 to 3 show the average classification accuracy over all datasets on the SVM, Naive Bayes, and AdaBoost classifiers, respectively. As can be seen in these figures, on SVM and AB classifiers, the PSO-based method had the highest average classification accuracy, and on the Naive Bayes classifier, COA-based method won the highest rank. The results of Figure 1 show that the PSO-based obtained 89.35 % average classification accuracy and achieved the first rank with a margin of 0.50 percent compared to the ACO-based method, which obtained the second-best average classification accuracy. Moreover, from the Figure 2 results, it can be seen that the differences between the obtained classification accuracy of the COA-based method and the second-best ones (PSO-based) and third-best ones (ACO-based) on Naive Bayes classifier were reported 0.16 (i.e., 89.24 – 89.08) and 0.88 (89.24-88.36) percent. Furthermore, on the AB classifier, the PSO-based feature selection method gained the first rank with an average classification accuracy of 88.92 %, and the ACO-based and ABC-based feature selection methods were ranked second and third with an average classification accuracy of 88.28 % and 87.73 %, respectively.

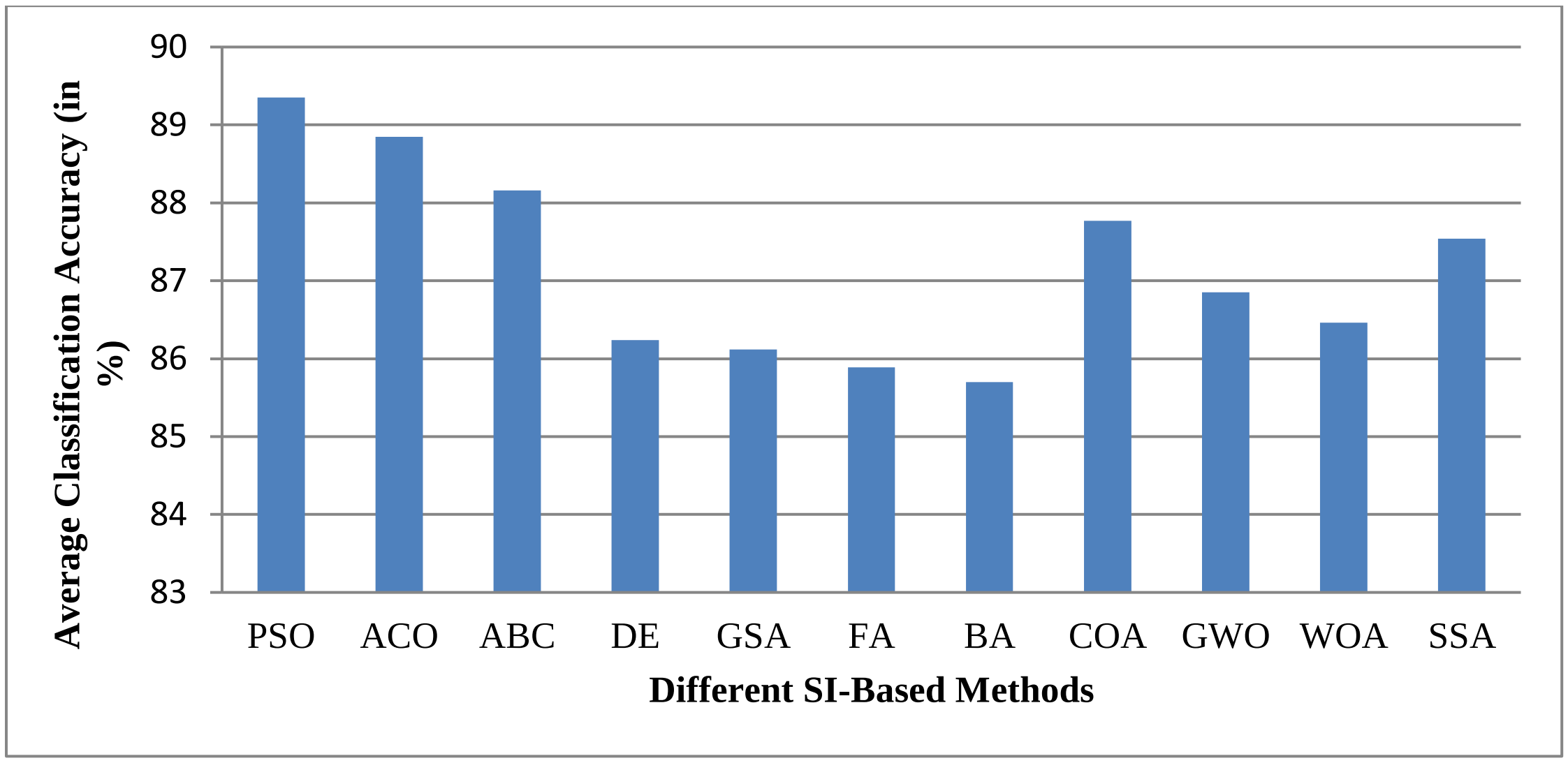

**Figure 1:** Average classification accuracy over all datasets on the SVM classifier (wrapper-based methods).

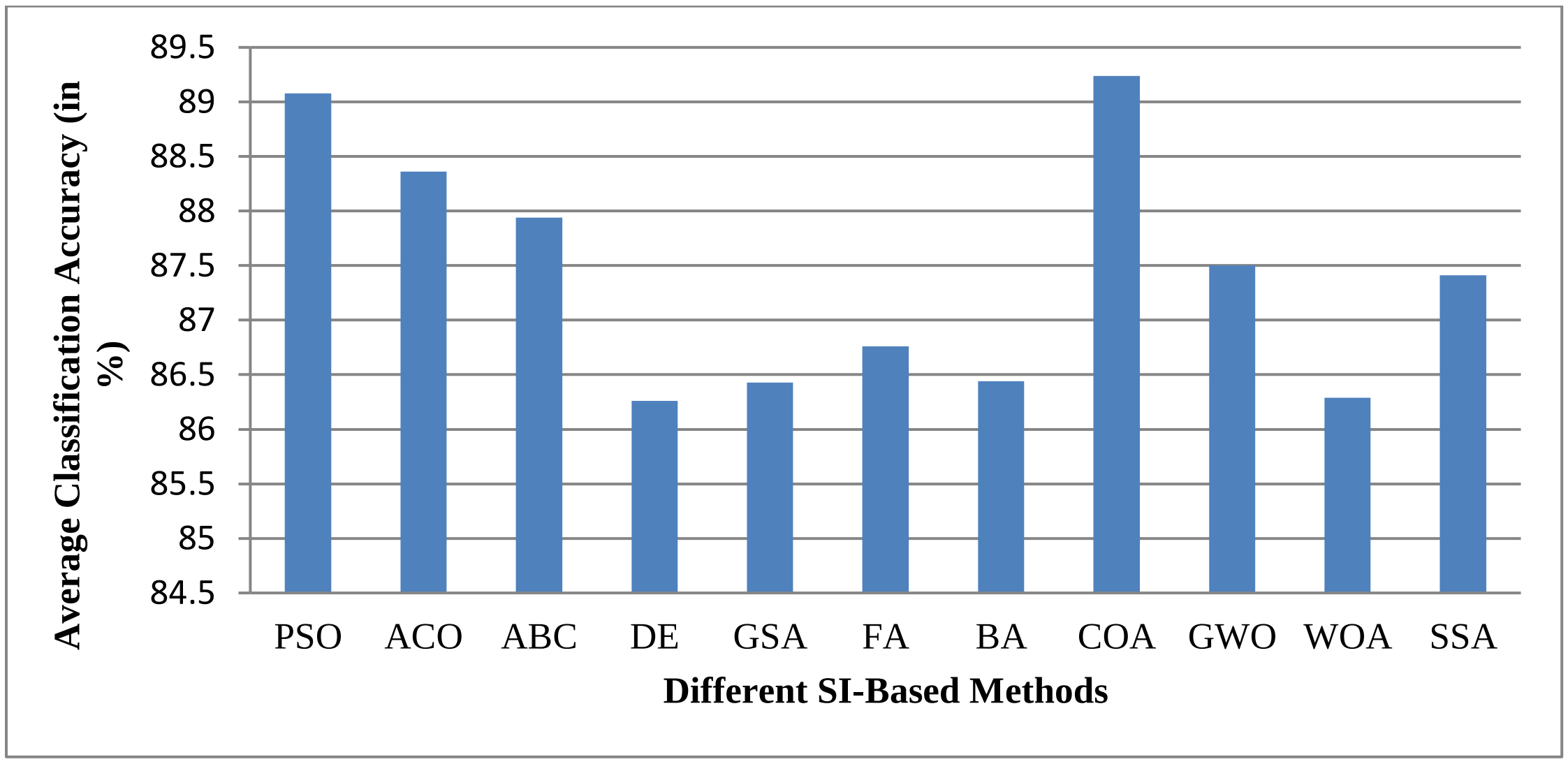

**Figure 2:** Average classification accuracy over all datasets on the Naive Bayes classifier (wrapper-based methods).

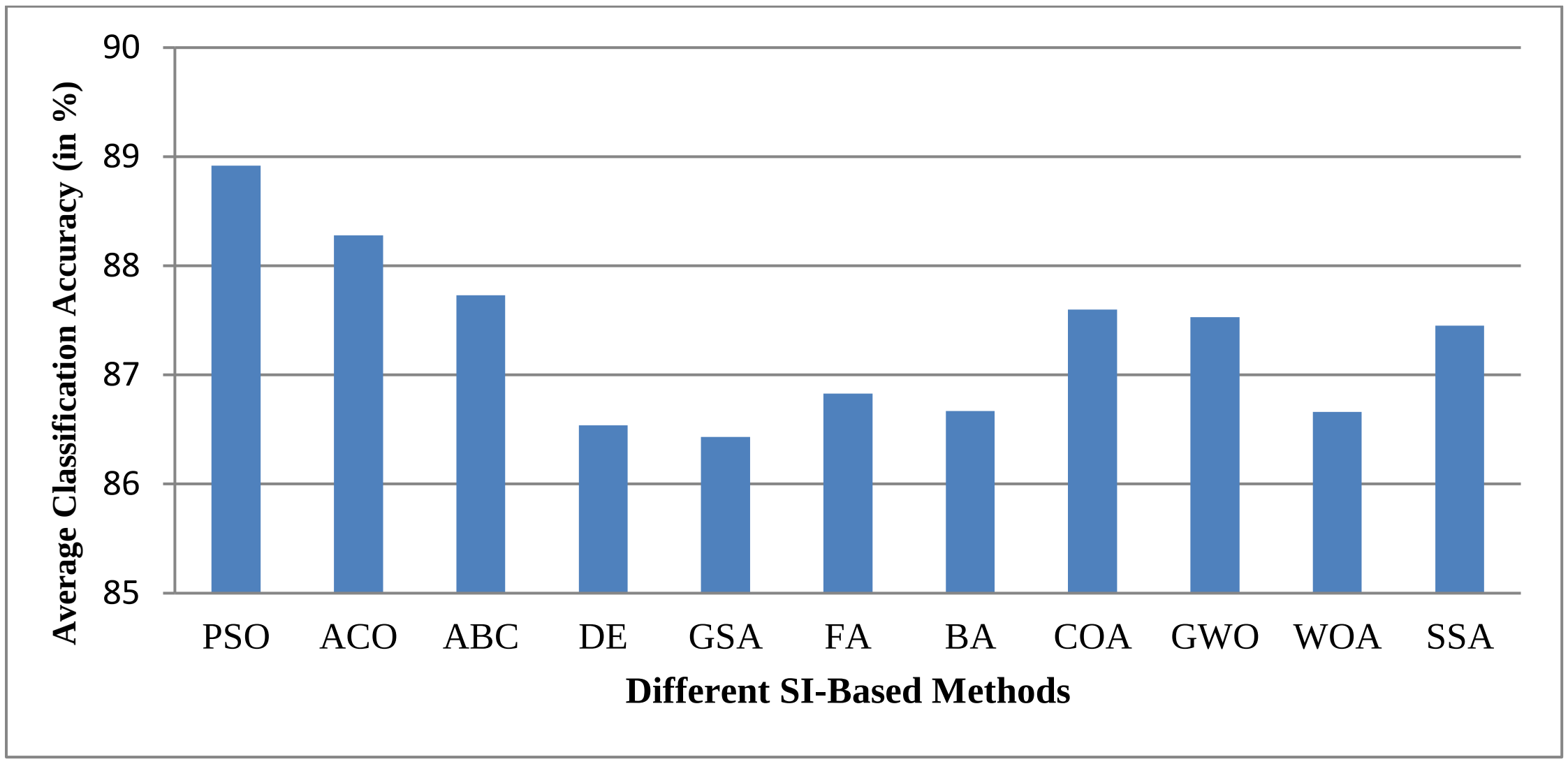

**Figure 3:** Average classification accuracy over all datasets on the AdaBoost classifier (wrapper-based methods).

Table 5 records the number of selected features of the eleven wrapper SI-based feature selection methods for each dataset. It can be observed that, generally, all the eleven methods achieve a significant reduction of dimensionality by selecting only a small portion of the original features. Among the various methods, in the Arrhythmia, Madelon, Isolet, and Colon datasets, the PSO-based method has the best performance among the other SI-based, selecting only 7.21, 14.87, 22.95, and 0.58 %, respectively. Moreover, in the SpamBase and Sonar datasets, the ACO-base method selected an average of 8.18 and 7.11 features, respectively.

**Table 5:** Average number of selected features of the different wrapper SI-based methods. (Minimum number of selected features is indicated in bold face and underlined and the second best is in bold face)

| Dataset | Number of the original feature | Method | Number of selected features | The ratio of the selected features to the original features (in %) |
|---|---|---|---|---|
| **SpamBase** | 57 | PSO | **8.91** | **15.63** |
| | | ACO | <u>**8.18**</u> | <u>**14.35**</u> |
| | | ABC | 8.98 | 15.75 |
| | | DE | 9.32 | 16.35 |
| | | GSA | 10.12 | 17.75 |
| | | FA | 10.03 | 17.60 |
| | | BA | 10.78 | 18.91 |
| | | COA | 9.64 | 16.91 |
| | | GWO | 10.73 | 18.82 |
| | | WOA | 11.62 | 20.39 |
| | | SSA | 9.17 | 16.09 |
| **Sonar** | 60 | PSO | 7.32 | 12.20 |
| | | ACO | <u>**7.11**</u> | <u>**11.85**</u> |
| | | ABC | 7.89 | 13.15 |
| | | DE | 8.19 | 13.65 |
| | | GSA | 8.28 | 13.80 |
| | | FA | 7.98 | 13.30 |
| | | BA | 8.08 | 13.47 |
| | | COA | **7.26** | **12.10** |
| | | GWO | 9.71 | 16.18 |
| | | WOA | 8.48 | 14.13 |
| | | SSA | 8.31 | 13.85 |
| **Arrhythmia** | 279 | PSO | **20.12** | **7.21** |
| | | ACO | 21.88 | 7.84 |
| | | ABC | 20.96 | 7.51 |
| | | DE | 23.14 | 8.29 |
| | | GSA | 22.61 | 8.1 |
| | | FA | 24.08 | 8.63 |
| | | BA | 22.83 | 8.18 |
| | | COA | 21.39 | 7.67 |
| | | GWO | 23.94 | 8.58 |
| | | WOA | 22.06 | 7.91 |
| | | SSA | 23.70 | 8.49 |
| **Madelon** | 500 | PSO | **74.35** | **14.87** |
| | | ACO | 68.91 | 13.78 |
| | | ABC | **75.19** | **15.04** |
| | | DE | 77.84 | 15.57 |
| | | GSA | 81.56 | 16.31 |
| | | FA | 82.93 | 16.59 |
| | | BA | 88.48 | 17.70 |
| | | COA | 82.38 | 16.48 |
| | | GWO | 79.19 | 15.84 |
| | | WOA | 85.73 | 17.15 |
| | | SSA | 78.03 | 15.61 |
| **Isolet** | 617 | PSO | **141.63** | **22.95** |
| | | ACO | **142.56** | **23.10** |
| | | ABC | 172.78 | 28.00 |
| | | DE | 163.98 | 26.58 |
| | | GSA | 159.62 | 25.87 |
| | | FA | 152.83 | 24.77 |
| | | BA | 143.48 | 23.25 |
| | | COA | 145.85 | 23.64 |
| | | GWO | 158.61 | 25.71 |
| | | WOA | 164.92 | 26.73 |
| | | SSA | 147.67 | 23.93 |
| **Colon** | 2000 | PSO | **11.63** | **0.58** |
| | | ACO | 12.98 | 0.64 |
| | | ABC | 13.26 | 0.66 |
| | | DE | 13.64 | 0.68 |
| | | GSA | 12.76 | 0.64 |
| | | FA | 12.29 | 0.61 |
| | | BA | 13.61 | 0.68 |
| | | COA | **12.06** | **0.60** |
| | | GWO | 14.98 | 0.75 |
| | | WOA | 13.21 | 0.66 |
| | | SSA | 12.45 | 0.62 |

Also, several experiments were conducted to compare the execution time of different wrapper SI-based feature selection methods. In these experiments, corresponding execution times (in second) for each method, were reported in Table 6. Due to the fact that the feature selection process and the final classification process are independent, only the execution time for feature selection is reported in the data in this Table. It can be seen from the results that generally the single objective SI-based feature selection methods are much faster than the multi objective SI-based feature selection methods (i.e., ACO-based and ABC-based). This is due to the fact that in multi objective methods, several different criteria are usually considered to calculate the fitness of solutions; thus, these methods can be computationally more expensive than the single objective methods. Moreover, the reported results revealed that the PSO-based feature selection method has the lowest average execution time over all dataset among all other methods. After the PSO-based method, WOA-based and GSA-based methods ranked second and third, respectively.

**Table 6:** Average execution time (in second) of wrapper feature selection methods over ten independent runs.

| Dataset | Wrapper-based feature selection method | | | | | | | | | | |
|---|---|---|---|---|---|---|---|---|---|---|---|
| | **PSO** | **ACO** | **ABC** | **DE** | **GSA** | **FA** | **BA** | **COA** | **GWO** | **WOA** | **SSA** |
| **SpamBase** | 6.78 | 8.41 | 8.93 | 9.44 | 6.82 | 7.68 | 7.83 | 9.12 | 8.73 | 7.11 | 8.64 |
| **Sonar** | 4.19 | 7.81 | 8.27 | 7.93 | 6.54 | 6.08 | 7.18 | 7.19 | 8.63 | 5.09 | 7.62 |
| **Arrhythmia** | 21.93 | 27.81 | 29.98 | 27.18 | 22.38 | 24.71 | 23.49 | 25.62 | 26.03 | 24.78 | 26.61 |
| **Madelon** | 89.51 | 98.32 | 108.67 | 109.67 | 99.32 | 101.56 | 105.84 | 99.53 | 104.86 | 88.73 | 101.78 |
| **Isolet** | 48.18 | 51.4 | 58.90 | 55.09 | 48.91 | 55.78 | 54.32 | 48.18 | 56.71 | 48.36 | 52.07 |
| **Colon** | 59.81 | 78.42 | 61.76 | 60.31 | 54.78 | 58.17 | 59.46 | 53.77 | 59.14 | 58.92 | 54.05 |
| **Average** | **38.4** | **45.36** | **46.09** | **44.94** | **39.79** | **42.33** | **43.02** | **40.57** | **44.02** | **38.83** | **41.8** |

All of the methods evaluated in the first part of this subsection were in the category of wrapper-based feature selection methods. In the reminder of this subsection, some SI-based methods that use filter approaches to search the final feature subset will be evaluated. These methods include PSO-based (Zhang, Ding et al. 2018), ACO-based (Moradi and Rostami 2015), DE-based (Hancer, Xue et al. 2018) and WOA-based (Nematzadeh, Enayatifar et al. 2019).

In the first experiments, the performances of different filter-based methods are evaluated over different classifiers. Tables 7 record the average classification accuracy (in %) over ten independent runs of the different filter SI-based feature selection methods using SVM, NB, and AB classifiers. Each entry of the Tables 7 shows the mean value as well as standard deviation (shown in parenthesis) of ten independent runs. The reported results of this Table shows that, in most cases, the ACO-based method performs better than the other filter SI-based feature selection method. For example, in colon dataset on the SVM classifier, ACO-based method obtained 84.53 % classification accuracy. In contrast, for PSO, DE and WOA-based method, these values were reported 81.42, 79.37, and 78.81, respectively.

**Table 7:** Average classification accuracy rate and as standard deviation (shown in parenthesis) over ten runs of the filter-based feature selection methods using SVM, Naive Bayes, and AdaBoost classifier. The best result is indicated in bold face and underlined, and the second-best is in bold face.

| **Dataset** | **Method** | **Classifier** | | |
|---|---|---|---|---|
| | | **SVM** | **Naive Bayes** | **AdaBoost** |
| **SpamBase** | PSO | 87.32 (1.83) | 86.51 (2.35) | **87.98 (3.24)** |
| | ACO | **88.54 (2.11)** | **87.67 (2.29)** | **88.20 (1.78)** |
| | DE | **88.13 (1.64)** | **88.01 (2.81)** | 87.62 (2.71) |
| | WOA | 87.06 (1.82) | 86.69 (2.95) | 86.17 (1.94) |
| **Sonar** | PSO | 86.03 (2.13) | **86.79 (2.43)** | **85.93 (3.72)** |
| | ACO | **87.74 (1.67)** | **86.28 (3.17)** | 85.28 (2.39) |
| | DE | **86.21 (2.28)** | 85.63 (2.38) | **85.89 (2.25)** |
| | WOA | 86.13 (3.15) | 86.12 (2.92) | 84.93 (2.59) |
| **Arrhythmia** | PSO | **61.32 (1.98)** | **61.84 (2.39)** | **60.95 (1.71)** |
| | ACO | 60.73 (2.93) | 60.08 (2.19) | **61.38 (2.48)** |
| | DE | **60.85 (4.86)** | **60.13 (2.36)** | 59.22 (2.70) |
| | WOA | 60.01 (2.72) | 59.94 (2.73) | 60.03 (1.94) |
| **Madelon** | PSO | **75.61 (2.84)** | **74.58 (2.16)** | **74.86 (2.66)** |
| | ACO | **75.32 (3.06)** | **75.12 (2.81)** | **74.93 (2.87)** |
| | DE | 73.28 (2.93) | 72.98 (3.09) | 73.38 (1.72) |
| | WOA | 73.71 (1.69) | 73.14 (2.62) | 72.86 (2.39) |
| **Isolet** | PSO | **81.23 (2.75)** | **80.72 (2.31)** | **81.43 (2.38)** |
| | ACO | 80.28 (2.94) | **80.46 (1.98)** | 80.98 (1.16) |
| | DE | **81.09 (2.98)** | 80.02 (2.72) | **81.16 (2.61)** |
| | WOA | 80.14 (3.11) | 79.81 (1.08) | 80.02 (1.09) |
| **Colon** | PSO | **81.42 (2.19)** | **80.98 (3.81)** | **81.38 (1.98)** |
| | ACO | **84.53 (1.92)** | **84.13 (2.63)** | **84.77 (2.13)** |

| | | | |
|---|---|---|---|
| DE | 79.37 (2.83) | 79.14 (3.17) | 79.06 (2.41) |
| WOA | 78.81 (1.08) | 79.56 (2.62) | 78.12 (1.81) |

Also, Figures 4 indicates the average classification accuracy over all datasets on the SVM, Naive Bayes, and AdaBoost classifiers. As can be seen in these reported results, on all classifiers, the ACO-based method had the highest average classification accuracy. For example, this figure shows that the ACO-based obtained 79.52 % average classification accuracy on SVM classifier and achieved the first rank with a margin of 0.70 percent compared to the PSO-based method, which obtained the second-best average classification accuracy. Furthermore, on the AB classifiers, the ACO-based feature selection method gained the first rank with an average classification accuracy of 79.26 %, and the PSO-based and DE-based feature selection methods were ranked second and third with an average classification accuracy of 78.76 % and 77.72 %, respectively.

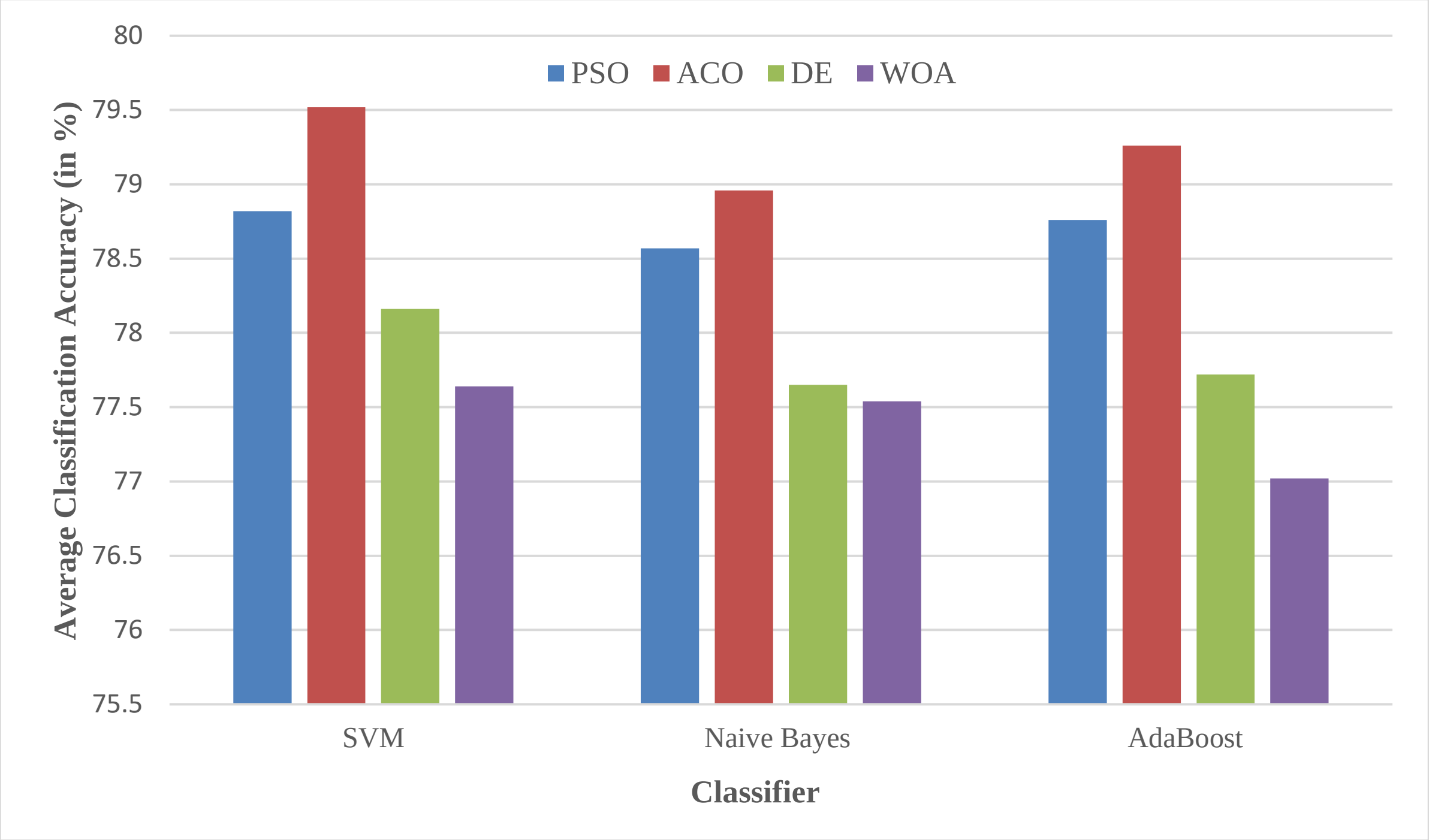

**Figure 4:** Average classification accuracy over all datasets on different classifiers (wrapper-based methods).

Moreover, in the Table 8 corresponding execution times (in second) for each filter SI-based feature selection method, were reported. Similar to previous execution time table, only the time of the feature selection process is considered as the execution time, in the Table 8. It can be seen that the multi objective wrapper feature selection method (i.e. DE-based method) has a higher execution time than other single objective methods. Moreover, the reported results revealed that the PSO-based feature selection method has the lowest average execution time over all dataset among all other methods. After the PSO-based method, ACO-based methods ranked second.

**Table 8:** Average execution time (in second) of filter feature selection methods over ten independent runs.

| Dataset | Filter-based feature selection method | | | |
|---|---|---|---|---|
| | **PSO** | **ACO** | **DE** | **WOA** |
| **SpamBase** | 3.47 | 4.19 | 4.97 | 4.32 |
| **Sonar** | 0.17 | 0.43 | 0.79 | 0.36 |
| **Arrhythmia** | 5.42 | 7.81 | 11.47 | 9.54 |
| **Madelon** | 11.34 | 14.87 | 12.38 | 12.98 |
| **Isolet** | 9.65 | 10.32 | 11.94 | 10.62 |
| **Colon** | 16.81 | 19.41 | 22.67 | 17.42 |
| **Average** | **7.81** | **9.50** | **10.70** | **9.20** |

## 4.5. Discussion

In this section, the strengths and weaknesses of the methods studied will be evaluated, and the factors that can lead to the superiority of a feature selection method will be analyzed.

1- Since each classifier has certain properties, the single classifiers are usually less accurate and generalization than the wrapper-based feature selection model that uses a combination of multiple classifiers. In other words, multiple-classifiers are usually highly accurate due to the diversity of used classifiers and the prediction performance of each single classifier. Unlike other compared wrapper SI-based feature selection methods that utilize a single classifier in their feature selection process to calculate the quality of a generated feature set, in the wrapper PSO-based feature selection method (Xue, Tang et al. 2020), four different classifiers include SVM, LDA, KNN, and ELM, are used as evaluation functions.

2- Given that in a large data set, the number of unrelated and redundant features has also increased, it is possible that evolutionary algorithms will be stuck in the local optimal. Moreover, many SI-based methods that use wrapper approaches to search for the optimal features subset are usually highly computationally complex and inefficient in high-dimensional datasets. Among the evaluated wrapper SI-based feature selection methods in this section, the PSO-based (Xue, Tang et al. 2020), ACO-based (Liu, Wang et al. 2019) COA-based (Elyasigomari, Lee et al. 2017), and SSA-based (Neggaz, Ewees et al. 2020) feature selection methods, highly accurate in the high-dimensional dataset (i.e., colon dataset with 2000 features), while other evaluated SI-based methods were only effective in low-dimensional data sets.

3- One of the main goals of an efficient feature selection method is to identify the optimal number of required features for the machine learning task and prevent the selection of too many or too few features during their feature selection process. If too many features are selected in a feature selection method, the probability of selecting redundant and irrelevant features will be increased; as a result, the prediction accuracy will be decreased. On the other hand, if too few features are selected, they will not be able to represent all the information of original features. Among the studied methods in this paper, multi-objective methods and methods that took into account the number of selected features in their fitness function showed better performance, and the number of final selected features by these methods were fewer.

4- Exploration of the search space and exploitation of the best solutions found are two conflicting objectives that must be taken into account when using a swarm intelligence-based method. Exploration means to generate diverse solutions so as to explore the search space on a global scale, while exploitation means to focus the search in a good solution region. A good balance between these two objectives will improve the performance of the searching method. Good SI-based feature selection methods should employ different strategies for their search processes, in order to develop a powerful method with a better balance between exploration and exploitation capabilities and better convergence speed. Among the studied wrapper methods in this paper, PSO-based (Xue, Tang et al. 2020), ACO-based (Liu, Wang et al. 2019), ABC-based, SSA-based (Neggaz, Ewees et al. 2020) method, demonstrated better performance in balancing the factors of exploration and exploitation. Furthermore, among the evaluated filter-based methods, the ACO-based method (Moradi and Rostami 2015) showed the best performance for the balance between exploration and exploitation.

5- A feature selection method can be evaluated from two aspects: efficiency and effectiveness. The efficiency of a feature selection method depends on the required time to find the final feature subset. While effectiveness depends on the quality of the selected feature subset. These two criteria have been in conflict with each other, and usually, the improvement of one of them leads to the reduction of the other. Therefore, balancing these two criteria is an important and necessary issue in feature selection. Wrapper-based feature selection methods, due to the use of the learning algorithm in the feature selection process, will be able to effectively select a feature subset of relevant and non-redundant features. Therefore, these wrapper-based methods are usually highly accurate. On the other, these methods have high computational complexity and will have a high execution time in high-dimensional datasets. Also, filter-based methods are much more efficient than wrapper-based methods in terms of computational complexity due to the lack of learning algorithms in the feature selection process. But most filter-based methods converge to local optimal, and the quality of the selected subset in this approach is usually less than the wrapper-based methods.

## 5. Conclusions

With the advancement of data collection technologies and the increasing capacity of data storage over the last decades, high-dimensional datasets have grown significantly. Usually, many features of these datasets are irrelevant or redundant, which reduces the performance of the prediction model. Feature selection plays an essential role in machine learning and, more specifically, in the high-dimensional dataset. Reducing the size of the medical dataset, on the one hand, reduces the computational complexity and, on the other hand, decrees the parameters of the classification algorithm. As a result, the accuracy of the prediction model will be increased. In the past decades, the rapid growth of computer and database technologies has led to the rapid growth of large-scale datasets. On the other hand, data mining applications with high dimensional datasets that require high speed and accuracy are rapidly increasing. An important issue with these applications is the curse of dimensionality, where the number of features is much greater than the number of patterns. One of the dimensionality reduction approaches is feature selection that can increase the accuracy of the data mining task and reduce its computational complexity. The feature selection method aims at selecting a subset of features with the lowest inner similarity and highest relevancy to the target class. It reduces the dimensionality of the data by eliminating irrelevant, redundant, or noisy data.

In this paper, comparative analysis and categorization of different feature selection methods are presented. Moreover, in this paper, wrapper and filter SI-based method (i.e., PSO, ACO, ABC. DE, GSA, FA, BA, COA, GWO, WOA, and SSA) and its application in feature selection are studied. Furthermore, the strengths and weaknesses of the different studied SI-based feature selection methods are evaluated, and the factors that can lead to the superiority of these methods are analyzed.